\documentclass{article}
\usepackage{ijcai23}

\usepackage{times}
\usepackage{soul}
\usepackage{url}
\usepackage[hidelinks]{hyperref}
\usepackage[utf8]{inputenc}
\usepackage[small]{caption}
\usepackage{subcaption}
\usepackage{graphicx}
\usepackage{booktabs}
\usepackage{algorithm}
\usepackage{algorithmic}
\usepackage[switch]{lineno}
\usepackage{multirow}
\usepackage{tabularx}
\usepackage{amsmath}

\usepackage{amsfonts}
\newtheorem{definition}{Definition}
\usepackage{mathtools}
\DeclareMathOperator*{\argmax}{arg\,max}

\title{A Unified View of Deep Learning for Reaction and Retrosynthesis Prediction: Current Status and Future Challenges}

\author{
Ziqiao Meng$^{1}$\thanks{This work is done when Ziqiao Meng worked as an intern in Tencent AI Lab.}
\and Peilin Zhao$^{2}$\thanks{Corresponding authors: Peilin Zhao, and Irwin King.}
\and Yang Yu$^{2}$
\And Irwin King$^{1}$\footnotemark[2]
\affiliations
$^1$The Chinese University of Hong Kong\\
$^2$Tencent AI Lab
\emails
\{zqmeng, king\}@cse.cuhk.edu.hk,
masonzhao@tencent.com
}
\begin{document}

\maketitle

\begin{abstract}
    Reaction and retrosynthesis prediction are fundamental tasks in computational chemistry that have recently garnered attention from both the machine learning and drug discovery communities. Various deep learning approaches have been proposed to tackle these problems, and some have achieved initial success. In this survey, we conduct a comprehensive investigation of advanced deep learning-based models for reaction and retrosynthesis prediction. We summarize the design mechanisms, strengths, and weaknesses of state-of-the-art approaches. Then, we discuss the limitations of current solutions and open challenges in the problem itself. Finally, we present promising directions to facilitate future research. To our knowledge, this paper is the first comprehensive and systematic survey that seeks to provide a unified understanding of reaction and retrosynthesis prediction.  
\end{abstract}

\section{Introduction}

Drug discovery is crucial to human healthcare, but the process is notoriously labor-intensive and costly. As Eroom's law suggests \cite{Eroom}, the exploration of new drugs becomes increasingly slower and more expensive over time. Therefore, it is natural and significant to leverage machine learning techniques to accelerate the drug discovery process. In recent years, the use of deep learning approaches to enhance different stages of drug discovery has become prevalent due to the rise of deep learning. Among these stages, reaction prediction and retrosynthesis prediction are two fundamental steps that can benefit from deep learning tools.

In actual production environments, chemists often aim to design synthesis routes that can lead to target molecules through a series of chemical reactions. One common strategy is to decompose target molecules into simpler precursor structures that can be synthesized more easily, a process known as retrosynthetic analysis. 
\begin{figure}[ht]
    \centering
    \includegraphics[width=0.45\textwidth]{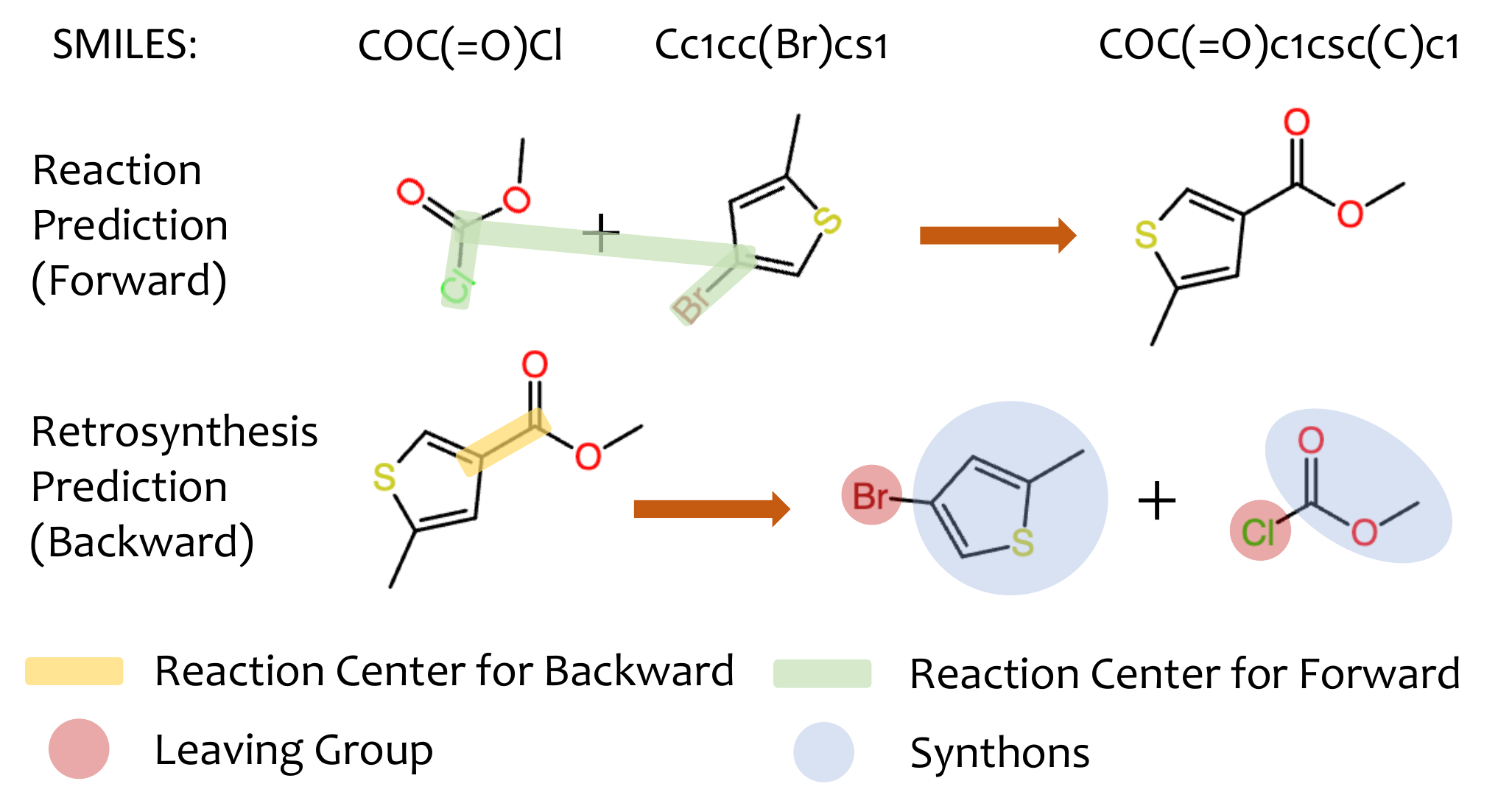}
    \caption{This figure illustrates the problem formulation of reaction and retrosynthesis prediction}
    \label{fig:problem-formulation}
\end{figure}
Automating this planning process using deep learning is crucial for discovering and optimizing synthesis routes. Retrosynthetic planning involves two subtasks: multi-step retrosynthetic planning and single-step retrosynthesis prediction. In this survey, we focus on the latter, as multi-step planning is usually approached as a searching problem, which is fundamentally different from the conditional structured prediction problem posed by reaction prediction and single-step retrosynthesis prediction. For the remainder of this survey, we will refer to single-step retrosynthesis prediction as simply retrosynthesis prediction. Reaction prediction is another key task in organic synthesis analysis. A robust reaction prediction model can provide insight into the underlying mechanisms of biochemical reactions, as well as generate virtual reactions to expand the database for retrosynthetic planning. In summary, reaction and retrosynthesis prediction are interrelated and can enhance one another.

Several surveys exist on reaction and retrosynthesis prediction. Engkvist \cite{reaction-survey} provides an overview of various computational approaches for reaction prediction, ranging from quantum computation to deep learning tools. However, this survey lacks detail for each approach and provides limited coverage of deep learning-based methods. On the other hand, \cite{retro-survey} reviews some deep learning-based retrosynthetic planning methods and datasets, but it falls behind the state-of-the-art research on retrosynthesis prediction. It is worth noting that DualTF \cite{RetroEBM} provides a unified framework for retrosynthesis prediction from the perspective of energy-based models, but it does not offer a unified understanding of reaction and retrosynthesis, nor does it provide specific discussions on the limitations and major challenges of each approach. Overall, the existing literature lacks a comprehensive and unified understanding of advanced reaction and retrosynthesis prediction models.

Compared to previously mentioned surveys and works, our survey unifies the formulation of reaction and retrosynthesis prediction for the first time. We systematically discuss the strengths and weaknesses of each approach from different perspectives for both problems. Additionally, our survey presents novel challenges and limitations that were not explicitly stated in previous works. Finally, based on the current status, we list several future directions for further enhancement, along with detailed analysis.

\section{Preliminary and Problem Formulation}
\label{section3}

Reaction prediction and retrosynthesis prediction are dual tasks of each other. They are also known as forward reaction prediction and backward reaction prediction, respectively. To apply deep learning, both tasks are formulated as conditional generation tasks. In this section, we first introduce the problem formulation for both tasks. Then, we provide an introduction to the basic background knowledge required for understanding different methods.
%we sequentially introduce the problem definitions and problem formulations of reaction prediction and retrosynthesis prediction. These two problems are dual problems to each other and therefore they are also known as forward reaction prediction and backward reaction prediction. 

\paragraph{Molecular Formulation.} A chemical molecule $\mathcal{M}$ can be represented in two major data formats: the \textit{SMILES string} and the \textit{molecular graph}. (1) For the SMILES format, a molecular structure $\mathcal{M}$ is described as a sequence of characters $m_{i}$ such that $\mathcal{M} \coloneqq m_{1}m_{2}...m_{L}$, where $L$ denotes the total length of the string. The sequence represents a spanning tree of the 2D molecular structure, and each character $m_{i}$ denotes a structural element such as an atom element, chemical bond, branching notation, and so on. (2) A molecule can also be abstracted as an undirected graph $\mathcal{G} = \lbrace \mathcal{V}, \mathcal{E}\rbrace$, where $\mathcal{V} = \lbrace v_{1},.., v_{n}\rbrace$ denotes the set of $n$ atoms and $\mathcal{E} = \lbrace e_{1},.., e_{m}\rbrace$ denotes the set of $m$ edges. Each node is associated with a feature vector $\mathbf{h}_{i}\in \mathbb{R}^{d}$ containing atomic information such as aromaticity and electric charge. Then, we have a feature matrix $\mathbf{H} \in \mathbb{R}^{n\times d}$ containing all-atom information. An adjacency matrix $\mathbf{A}\in \mathbb{R}^{n\times n\times c}$ describes the topological structures of $\mathcal{M}$, where $\mathbf{A}_{ijk}$ indicating the presence or absence of a chemical bond of type $k$ between atom $i$ and atom $j$. Multiple molecules can be easily represented by the above two formats. For SMILES format, multiple SMILES strings can be concatenated by a full stop “.” into one single SMILES sequence. For molecular graph, a set of molecules is regarded as a single disconnected graph with each molecule as an independent connected component.  
\begin{definition} 
(Reaction Prediction) Given a set of $N$ reactant molecules $\lbrace \mathcal{M}^{R}_{i}\rbrace_{i=1}^{N}$, the target is to predict the set of $M$ possible product molecules $\lbrace \mathcal{M}^{P}_{i}\rbrace_{i=1}^{M}$. 
\end{definition}

\begin{definition}
(Retrosynthesis Prediction) Given a set of $M$ product molecules $\lbrace \mathcal{M}^{P}_{i}\rbrace_{i=1}^{M}$, the target is to predict a set of $N$ reactant molecules $\lbrace \mathcal{M}^{R}_{i}\rbrace_{i=1}^{N}$ that can lead to $\lbrace \mathcal{M}^{P}_{i}\rbrace_{i=1}^{M}$. 
\end{definition}
Note that in real implementations, $M=1$ because only the main product is recorded in public benchmark datasets. Unfortunately, this issue makes retrosynthesis prediction more difficult than reaction prediction because $M<N$ requires retrosynthesis to attach newly appeared atoms, leading to a much larger combinatorial search space. In general, reaction prediction aims to model the conditional probability distribution $\mathcal{P}(\mathcal{M}^{P}|\lbrace \mathcal{M}^{R}_{i}\rbrace_{i=1}^{N})$, while retrosynthesis prediction aims to model the distribution $\mathcal{P}(\lbrace \mathcal{M}^{R}_{i} \rbrace_{i=1}^{N}|\mathcal{M}^{P})$.
%\paragraph{\textbf{Single Step.}} People might ask a question that why we only consider the single case of product. First reason is following conventions of all related works. Second reason is that multiple products case is just a combination of single product cases. For example, if the retrosynthesis model could successfully make predictions in single product case, then we only need to apply the model individually on each product in multiple products case. 
\paragraph{Reaction Center \& Reaction Template.} In reaction prediction, a reaction center $\mathcal{C}$ is a subset of atom pairs $\mathcal{C} = \lbrace (v_{i}, v_{j})\rbrace\subseteq \mathcal{V}\times \mathcal{V}$ that change bond types when a chemical reaction occurs. In retrosynthesis prediction, a reaction center $\mathcal{C}$ is defined as a subset of existing bonds $\mathcal{C} = \lbrace e_{i}\rbrace\subseteq \mathcal{E}$ that can be modified to obtain simpler structures. A reaction template pool $\mathcal{T}$ is a set of reaction subgraph rules derived from a large chemical reaction database. A reaction template $T \in \mathcal{T}$ is an extracted subgraph pattern from the corresponding reaction center $\mathcal{C}$. Specifically, $T \coloneqq t^{R}_{1} + t^{R}_{2} + ... t^{R}_{N} \rightarrow t^{P}$, where $t^{R}_{i}$ denotes the subgraph pattern inside the $i$th reactant, and $t^{P}$ denotes the subgraph pattern inside the product. 

\paragraph{Atom-mapping.} Both reaction and retrosynthesis prediction follow the atom-mapping principle. This principle states that each atom in the reactants/products has exactly one corresponding atom in the products/reactants. This fundamental one-to-one mapping relation physically constrains reaction space and determines that chemical reactions are mainly about bond breaking and bond formation.

\paragraph{Synthon and Leaving Group.} A target molecule can be decomposed into a set of synthons $\mathcal{S} = \lbrace \mathcal{G}^{S}_{i}\rbrace_{i=1}^{N}$, which are simpler precursor substructures that can constitute the target molecule with few additional bond connections. Note that the atom set $\mathcal{V}$ covered by $\mathcal{S}$ is exactly the same as the target molecule. Leaving group $\mathcal{L} = \lbrace \mathcal{G}^{L}_{i}\rbrace_{i=1}^{N}$ is a group of atoms or substructures in original reactants that do not appear in the target molecule after a reaction occurs. In short, synthon $\mathcal{S}$ and leaving group $\mathcal{L}$ can form reactants $\lbrace \mathcal{G}^{R}_{i}\rbrace_{i=1}^{N}$. 

\paragraph{Evaluation Metrics.} Both reaction and retrosynthesis prediction adopt top-$k$ accuracies to evaluate the model performance. Top-$k$ accuracy is the percentage of reactions that have the ground-truth product in the top-$k$ predicted sets of molecules. As long as the top-$k$ predicted products include the ground-truth main product, it would be counted as a correct prediction. Usually, the $k$ is in range $\lbrace 1,2,3,5,10 \rbrace$. 

\section{Deep Learning on Reaction and Retrosynthesis Prediction}
\label{section4}

In this section, we discuss different approaches by classifying them into four categories which are template-based methods, sequence-based and graph-based autoregressive models, graph-based two-stage models, and graph-based non-autoregressive models. Specifically, we introduce their learning mechanism design decisions, weaknesses, and strengths. 

\begin{figure*}[htp]
\centering
\includegraphics[width=0.86\textwidth]{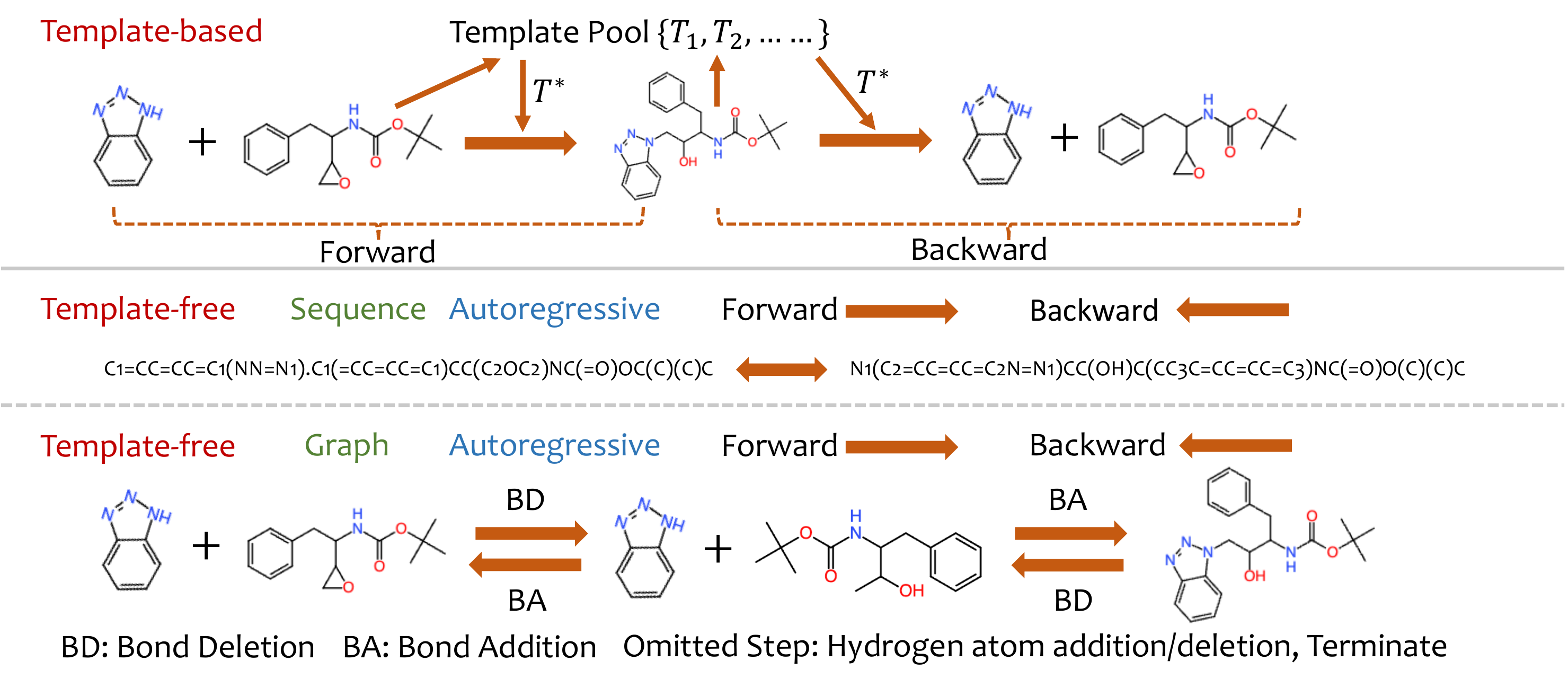}
\caption{This figure illustrates template-based models, sequence-based autoregressive models, and graph-based autoregressive models.} 
\label{fig:method-1}
\end{figure*} 

\subsection{Template-based Methods}

Template-based (TB) methods are mainly leveraging a pool of reaction templates to deduce possible reaction centers. TB methods are mimicking how human experts conduct chemical reasoning. Assume we have a pool of reaction templates $\mathcal{T} = \lbrace T_{1}, T_{2}, ...\rbrace$, then the inference process of TB methods aims at picking the best template as follows: 
\begin{equation}
\begin{split}
\label{eq:TB-reaction}
    & T^{*} = \argmax_{T_{i}\in \mathcal{T}}\mathcal{P}(T=T_{i}|\lbrace \mathcal{M}^{R}_{i}\rbrace_{i=1}^{N}), \\
    & \hat{\mathcal{M}}^{P} = \argmax_{\mathcal{M}_{i}^{P}\in M^{P}}\mathcal{P}(\mathcal{M}^{P}=\mathcal{M}_{i}^{P}|T^{*}), 
\end{split}
\end{equation} 
where $M^{P}$ is a pool of candidate product molecules. Eq.~(\ref{eq:TB-reaction}) describes the steps of choosing the best template $T^{*}$ and deducing the possible product $\hat{\mathcal{M}}^{P}$ according to the given $T^{*}$. The critical part is matching molecules with templates based on some similarity measure. Specifically, $\mathcal{P}(T_{i}|\lbrace \mathcal{M}^{R}_{i}\rbrace_{i=1}^{N})$ and $\mathcal{P}(\mathcal{M}^{P}_{i}|T^{*})$ are evaluated in the following way:
\begin{equation}
\begin{split}
    & \mathcal{P}(T_{i}|\lbrace \mathcal{M}^{R}_{i}\rbrace_{i=1}^{N}) = \frac{\exp(\mathrm{sim}(T_{i}, \lbrace \mathcal{M}^{R}_{i}\rbrace_{i=1}^{N}))}{\sum_{T_{j}\in \mathcal{T}}\exp(\mathrm{sim}(T_{j}, \lbrace \mathcal{M}^{R}_{i}\rbrace_{i=1}^{N})))}, \\
    & \mathcal{P}(\mathcal{M}_{i}^{P}|T^{*}) = \frac{\exp(\mathrm{sim}(T^{*}, \mathcal{M}_{i}^{P}))}{\sum_{\mathcal{M}_{j}^{P}\in M^{P}}\exp(\mathrm{sim}(T^{*}, \mathcal{M}_{j}^{P})))}, 
\end{split}
\end{equation}
where $\mathrm{sim}(\cdot,\cdot)$ denotes a similarity measure function. It can be a simple inner product between molecular embedding and template embedding obtained from deep neural networks, or a more complicated similarity function evaluated by some subgraph matching algorithms. TB retrosynthesis has the same learning mechanism but in a reverse direction: 
\begin{equation}
\begin{split}
\label{eq:TB-similar}
    &T^{*} = \argmax_{T_{i}\in \mathcal{T}}\mathcal{P}(T=T_{i}|\mathcal{M}^{P}),\\ 
    &\mathcal{P}(\mathcal{M}^{R}|T^{*}) = \prod_{i=1}^{N}\mathcal{P}(\mathcal{M}_{i}^{R}|t_{i}^{R}),\\
    &\hat{\mathcal{M}}_{i}^{R} = \argmax_{\mathcal{M}_{j}^{R}\in M^{R}}\mathcal{P}(\mathcal{M}^{R}=\mathcal{M}_{j}^{R}|t^{R}_{i}),\\
\end{split}
\end{equation}
where $M^{R}$ denotes the pool of candidate reactant molecules, $T^{*} \coloneqq t^{R}_{1} + t^{R}_{2} + ... t^{R}_{N}\rightarrow t^{P}$. The second and third equation in Eq.~(\ref{eq:TB-similar}) indicates that each predicted reactant molecule $\mathcal{\hat{M}}_{i}^{R}$ matches with the subgraph template $t_{i}^{R}$ with highest score. 

\paragraph{Advantages.} (1) TB methods are reliable since they are using extracted human knowledge, which can always provide good interpretations for predictions. (2) The training and inference process of TB methods are relatively simple, which is easy for domain experts to manipulate. 
\paragraph{Disadvantages.} (1) The performance of TB methods highly relies on the scale of template database. Therefore, the template database must be updated frequently, which is apparently very expensive. (2) TB methods have poor generalization to out-of-domain unseen reactions. (3) Templates are extracted local subgraph rules while ignoring global information over the reaction. Consequently, TB methods fail to capture global information interactions and easily make false predictions based on local rules. The illustration of TB methods is shown in Figure~\ref{fig:method-1}. 

\begin{figure*}[htp]
\centering
\includegraphics[width=0.86\textwidth]{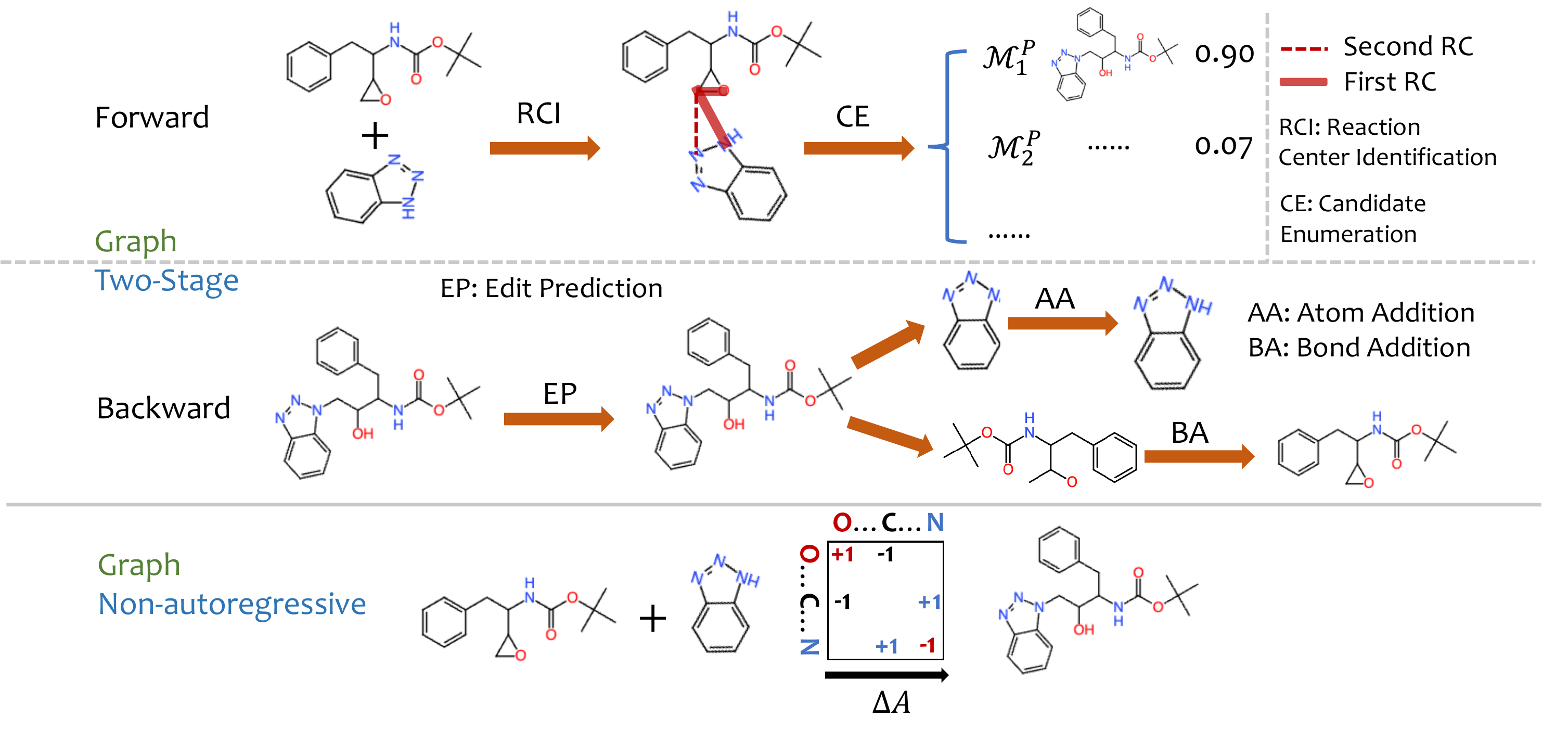}
\caption{This figure illustrates graph-based two-stage models and graph-based non-autoregressive models.} 
\label{fig:method-2}
\end{figure*} 

\subsection{Sequence-based and Graph-based Autoregressive Models}

Sequence-based autoregressive (SAR) models are widely adopted in both forward and backward prediction. It regards both problems as the neural machine translation problem. For reaction prediction, the input is SMILES strings of reactants $\mathcal{M}^{R} \coloneqq m^{R}_{1}m^{R}_{2}...m^{R}_{L_{1}}$ with length $L_{1}$ and output is SMILES strings of products $\mathcal{M}^{P} \coloneqq m^{P}_{1}m^{P}_{2}...m^{P}_{L_{2}}$ with length $L_{2}$. The input source and output target are flipped for retrosynthesis prediction. Specifically, SAR models are estimating the following conditional probability distribution:
\begin{equation}
\begin{split}
\label{eq:SAR}
    & \mathcal{P}(\mathcal{M}^{P}) = \prod_{i=1}^{L_{2}}\mathcal{P}(m^{P}_{i}|m^{P}_{<i}, \mathcal{M}^{R}), \\
    & \mathcal{P}(\mathcal{M}^{R}) = \prod_{i=1}^{L_{1}}\mathcal{P}(m^{R}_{i}|m^{R}_{<i}, \mathcal{M}^{P}),
\end{split}
\end{equation}
where $\mathcal{P}(m^{P}_{i}|m^{P}_{<i}, \mathcal{M}^{R})$ and $\mathcal{P}(m^{R}_{i}|m^{R}_{<i}, \mathcal{M}^{P})$ are approximated by the transformer \cite{transformer} model. Each generation step does a greedy search over token space and select the optimal token. To generate top-$k$ candidates, we only need to conduct beam search over greedy results. Graph-based autoregressive (GAR) models have similar learning mechanisms but with different generation sequence definitions. GAR models first define an action space $\mathcal{\pi}$ including several edit actions like atom addition/deletion, bond addition/deletion, and termination. A sequence of these actions would transform reactants/products into their corresponding products/reactants. Thus, they are estimating the following probability distribution:
\begin{equation}
\label{eq:GAR}
\begin{split}
    & \mathcal{P}(\mathcal{G}^{P}) = \prod_{i=1}^{L_{2}}\mathcal{P}(\pi_{i}|\pi_{<i}, \mathcal{G}^{R,<i}), \\
    & \mathcal{P}(\mathcal{G}^{R}) = \prod_{i=1}^{L_{1}}\mathcal{P}(\pi_{i}|\pi_{<i}, \mathcal{G}^{P,<i}),
\end{split}
\end{equation}
where $\mathcal{G}^{R,<i}$ and $\mathcal{G}^{P,<i}$ denote the states of edited reactants and products in previous $i-1$ steps respectively. The sampling process is exactly the same as that of SAR models. Each step $i$ selects the optimal action $\pi^{*}_{i}$ and applies the optimal action to transform reactants $\mathcal{G}^{R,i-1}$ and products $\mathcal{G}^{P,i-1}$ to next state $\mathcal{G}^{R,i}$ and $\mathcal{G}^{P,i}$ respectively. The optimal action would be “terminate” when all generation steps are predicted to be over. Note that both Eq.~(\ref{eq:SAR}) and Eq.~(\ref{eq:GAR}) can be simplified by Markov assumptions.

\paragraph{Advantages.} (1) Autoregressive models do not need atom-mapping information. (2) Autoregressive modeling has very natural sampling process, such as beam search. (3) Sequence-based modeling can directly utilize some mature techniques from natural language processing. 
\paragraph{Disadvantages.} (1) Autoregressive modeling can only generate predictions step-by-step, which is very inefficient. (2) Autoregressive modeling requires pre-defined generation orders. However, the molecular generation order is ambiguous. (3) Sequence-based modeling requires data augmentation techniques to improve performance. The illustration of both SAR models and GAR models are shown in Figure~\ref{fig:method-1}.

\subsection{Graph-based Two-Stage Models}

\paragraph{Reaction.} For graph-based two-stage models in reaction prediction, they split reaction prediction into two stages which are reaction center identification stage and candidate ranking stage. For reaction center identification, it aims at selecting atom pairs assigned with high reactivity scores: 
\begin{equation}
    \hat{\mathcal{C}} = \text{top-}{k}(\lbrace s(v_{i},v_{j}|\lbrace \mathcal{G}^{R}_{i}\rbrace_{i=1}^{N})\rbrace_{ij}), 
\end{equation}
where $s(\cdot)$ denotes any function that outputs a score in range $(0,1)$. Based on identified reaction centers $\hat{\mathcal{C}}$, a set of candidate products $\hat{G}^{P} = \lbrace \hat{\mathcal{G}}^{P}_{1},\hat{\mathcal{G}}^{P}_{2},...\rbrace$ will be enumerated in a combinatorial way through hand-written rules or reaction templates. Then the second stage is learning to rank candidate products in $\hat{G}^{P}$. To rank generated products, each pair $(\lbrace\mathcal{G}^{R}_{i}\rbrace_{i=1}^{N}, \hat{\mathcal{G}}^{P}_{i})$ should have a score evaluated as follows:
\begin{equation} 
    s(\lbrace \mathcal{G}^{R}_{i}\rbrace_{i=1}^{N}, \hat{\mathcal{G}}^{P}_{i}) = \sigma(f(\lbrace \mathcal{G}^{R}_{i}\rbrace_{i=1}^{N}, \hat{\mathcal{G}}_{i}^{P})), 
\end{equation}
%\frac{\exp(\mathrm{score}(\lbrace \mathcal{G}^{R}_{i}\rbrace_{i=1}^{N}, \hat{\mathcal{G}}_{i}^{P}))}{\sum_{\hat{\mathcal{G}}_{i}^{P}\in \hat{G}^{P}}\exp(\mathrm{score}(\lbrace \mathcal{G}^{R}_{i}\rbrace_{i=1}^{N}, \hat{\mathcal{G}}_{i}^{P})))}, 
where $f(\cdot, \cdot)$ can be complex neural networks and $\sigma(\cdot)$ denotes the sigmoid function. 

\paragraph{Retrosynthesis.} For graph-based two-stage models in retrosynthesis prediction, they split retrosynthesis into two stages which are edit prediction stage and synthon completion stage. For edit prediction, they select some existing edges with high predicted reactivity scores to decompose: 
\begin{equation}
    \hat{\mathcal{C}} = \text{top-}{k}(\lbrace s(e^{P}_{i})\rbrace_{i=1}^{m}),
\end{equation}
%\argmax_{\mathcal{C}\subseteq \mathcal{E}^{P}}\mathcal{P}(\mathcal{E}^{P}|\mathcal{G}^{P}),
After obtaining predicted edit centers $\hat{\mathcal{C}}$, we decompose the target molecule by breaking the predicted edit centers, which will results in a set of synthons $\mathcal{\hat{S}} = \lbrace \mathcal{\hat{G}}^{S}_{i}\rbrace_{i=1}^{N}$, ($N=k+1$). A basic assumption for this approach is that the number of synthons is the same as the number of reactants. For synthon completion stage, it models the following distribution: 
\begin{equation}
\label{eq:retro-stage2}
    \mathcal{P}(\mathcal{G}_{i}^{R}|\mathcal{\hat{G}}_{i}^{S}) = \mathcal{P}(\mathcal{G}_{i}^{L}|\mathcal{\hat{G}}_{i}^{S}). 
\end{equation}
The above Eq.~(\ref{eq:retro-stage2}) indicates that the synthon completion is equivalent to attaching the leaving group $\mathcal{G}_{i}^{L}$ to the corresponding synthon $\mathcal{G}_{i}^{S}$. Leaving group attachment can be either formulated as an autoregressive conditional generation problem or a classification problem with a predefined dictionary. With predicted leaving group $\hat{\mathcal{G}}_{i}^{L}$, each predicted reactant $\hat{\mathcal{G}}_{i}^{R}$ can be recovered by attaching $\hat{\mathcal{G}}_{i}^{L}$ to the corresponding synthon $\hat{\mathcal{G}}_{i}^{S}$. 

\paragraph{Advantages.} (1) The inference process of graph-based two-stage models is very similar to the deduction process of chemists. (2) Graph-based two-stage models split a hard task into two simpler tasks that are easier to tackle. \paragraph{Disadvantages.} (1) Some two-stage models like WLDN \cite{WLDN} require costly hand-crafted combinatorial enumerations. (2) The overall performance of two-stage models is restricted by the bottleneck of each stage. The illustration of the above methods are shown in Figure~\ref{fig:method-2}.

\subsection{Graph-based Non-autoregressive Models}

Only reaction prediction has a graph-based non-autoregressive model achieved by NERF \cite{NERF} and currently no non-autoregressive models for retrosynthesis. NERF reformulates the problem into electron redistribution modeling. Previous approaches adopt three dimensional adjacency matrix $\mathbf{A}\in \mathbb{R}^{n\times n\times c}$ while NERF adopts two-dimensional adjacency matrix by converting one-hot type encoding to scalar values such that $\mathbf{A}\in \mathbb{R}^{n\times n}$. Formally, $\mathbf{A}_{ij}$ is a scalar value in range $[0,3]$, representing the number of shared electrons (number of bonds) between atom $i$ and atom $j$. Note that aromatic bond is represented as $\mathbf{A}_{ij}=1$ and atoms $i$, $j$ will be marked as aromatic atoms. The key idea of NERF is predicting $\Delta \hat{\mathbf{A}}$ by combining self-attention mappings. NERF adopts the conditional variational autoencoder \cite{CVAE} (CVAE) architecture to approximate $\mathcal{P}(\mathcal{G}^{P}|\mathcal{G}^{R})$ by introducing a latent variable $z$. Instead of directly maximizing the log-likelihood $\log \mathcal{P}(\mathcal{G}^{P}|\mathcal{G}^{R})$, CVAE maximize its evidence-lower bound (ELBO):
\begin{equation}
\begin{split}
\label{eq:NERF}
   \log \mathcal{P}(\mathcal{G}^{P}|\mathcal{G}^{R})\geq &\mathbb{E}_{q(z|\mathcal{G}^{P},\mathcal{G}^{R})}[\log \mathcal{P}(\mathcal{G}^{P}|\mathcal{G}^{R},z)] -\\ 
    &\hspace{-0.2in}\text{KL}(q(z|\mathcal{G}^{P},\mathcal{G}^{R})||\mathcal{P}(z|\mathcal{G}^{R})),
\end{split}
\end{equation}
where $q(z|\mathcal{G}^{P},\mathcal{G}^{R})$ is the reaction encoder with reaction $(\mathcal{G}^{R},\mathcal{G}^{P})$ as input and low-dimensional representation $\mathbf{h}^{z}$ as output, $\mathcal{P}(\mathcal{G}^{P}|\mathcal{G}^{R},z)$ is product decoder with reactants $\mathcal{G}^{R}$ and latent embedding $\mathbf{h}^{z}$ as input, $\mathcal{P}(z|\mathcal{G}^{R})$ denotes the prior distribution of latent variable $z$. $\text{KL}$ term is minimizing the gap between $q(z|\mathcal{G}^{P},\mathcal{G}^{R})$ and $\mathcal{P}(z|\mathcal{G}^{R})$. The backbone network architectures of $q(z|\mathcal{G}^{P},\mathcal{G}^{R})$ are combinations of GNNs and transformers. With this architecture, $\mathcal{G}^{R}$ and $\mathcal{G}^{P}$ will be projected to reactants embedding $\mathbf{h}^{R}$ and products embedding $\mathbf{h}^{P}$ respectively. A cross attention layer is used for mapping $\mathbf{h}^{R}$ to latent $\mathbf{h}^{z}$ with $\mathbf{h}^{P}$ as teacher forcing during training. The conditional latent embedding is derived such that $\mathbf{\hat{h}}^{z}=\mathbf{h}^{R}+\mathbf{h}^{z}$. Then apply self-attention mechanism on $\mathbf{\hat{h}}^{z}$ to derive two electron redistribution matrix $W^{+}$ and $W^{-}$ for bond increase and bond decrease respectively. Then $\Delta \hat{\mathbf{A}} = W^{+} - W^{-}$ and $\mathcal{\hat{\mathbf{A}}^{P}} = \mathbf{A}^{R} + \Delta \hat{\mathbf{A}}$. 

\paragraph{Advantages.} (1) Non-autoregressive models enable parallel sampling, which result in much faster sampling speed compared to autoregressive ones. (2) Non-autoregressive models have already achieved the best class of top-1 accuracy, which demonstrates that the non-autoregressive decoder is very powerful for reaction modeling. (3) Non-autoregressive models do not need pre-defined generation order. \paragraph{Disadvantages.} (1) The uncertainty modeling of non-autoregressive models is very tricky. The top-k sampling process is not as natural as beam search in autoregressive modeling. (2) Non-autoregressive models rely on the atom-mapping information while this also requires additional algorithms for alignment. The illustration of NERF learning mechanism is shown in Figure~\ref{fig:method-2}. 

\section{Limitations and Challenges} 

\begin{table*}[ht]
    \fontsize{9}{10}\selectfont
    \setlength\tabcolsep{9pt}
    \centering
    \vskip 0.1in
    \resizebox{0.90\linewidth}{!}{%
    \begin{tabular}{c|c|c|c|c|c|c|c}
    \hline
       & Template Usage & Approaches & End-to-End & Generation & Graph/Sequence & Top-1 & Top-5 \\
       \hline
       \multirow{16}{5em}{Reaction Prediction (Forward)} & \multirow{4}{5em}{Template-based} & NN-reaction \cite{NN-reaction} & $\times$ & NA & Sequence & NR & NR\\
       & & NeuralSym \cite{neuralsym} & $\times$ & NA & Sequence & NR & NR \\
       & & Symbolic \cite{Symbolic}  & $\times$ & NA & Sequence & 90.4 & 95.0\\
       & & LocalTransform \cite{LocalTransform} & $\times$ & NA & Graph & 90.8 & 96.3 \\
       \cline{2-6}
       & \multirow{13}{5em}{Template-free} & WLDN \cite{WLDN} & $\times$ & NA & Graph & 79.6 & 89.2 \\
       & & GTPN \cite{GTPN} & $\checkmark$ & A & Graph & 83.2 & 86.5 \\
       & & MEGAN \cite{MEGAN} & $\checkmark$ & A & Graph & 89.3 & 95.6\\
       & & ELECTRO \cite{Electro} & $\checkmark$ & A & Graph & 77.8 & 94.7 \\
       & & Motif-Reaction \cite{motif-reaction} & $\checkmark$ & A & Sequence & 91.0 & 95.7\\
       & & MT-base \cite{moleculartransformer} & $\checkmark$ & A & Sequence & 88.8 & 94.4\\
       & & MT \cite{moleculartransformer} & $\checkmark$ & A & Sequence & 90.4 & 95.3 \\
       & & Graph2SMILES \cite{Graph2smiles} & $\checkmark$ & A & Sequence & 90.3 & 94.8 \\
       & & Chemformer \cite{Chemformer} & $\checkmark$ & A & Sequence & 91.3 & 93.7 \\
       & & Stransformer \cite{stransformer} & $\times$ & A & Graph & NR & NR\\
       & & ReactionT5 \cite{ReactionT5} & $\checkmark$ & A & Sequence & 88.9 & 95.2 \\
       & & Aug-Transformer \cite{Tetko} & $\checkmark$ & A & Sequence & 90.6 & 96.1 \\
       & & NERF \cite{NERF} & $\checkmark$ & NA & Graph & 90.7 & 93.7 \\
       & & ReactionSink \cite{reactionsink} & $\checkmark$ & NA & Graph & 91.3 & 94.0 \\
       \hline
       \multirow{19}{6em}{Retrosynthesis Prediction (Backward)} & \multirow{6}{5em}{Template-based} & RetroSim \cite{retrosim} & $\times$ & NA & Sequence & 52.9/37.3 & 81.2/63.3 \\
       & & RetroComposer \cite{Retrocomposer} & $\times$ & A & Graph & 65.9/54.5 & 89.5/83.2\\
       & & NeuralSym \cite{neuralsym} & $\times$ & NA & Graph & 55.3/44.4 & 81.4/72.4\\
       & & GLN \cite{GLN} & $\times$ & NA & Graph & 64.2/52.5 & 85.2/75.6 \\
       & & LocalRetro \cite{localretro} & $\times$ & A & Graph & 63.9/53.4 & 92.4/85.9\\
       & & DualTB \cite{RetroEBM} & $\times$ & NA & Sequence & 67.7/55.2 & 88.9/80.5\\
       \cline{2-6}
       & \multirow{13}{5em}{Template-free} & MEGAN \cite{MEGAN} & $\checkmark$ & A & Graph & 60.7/48.1 & 87.5/78.4\\
       & & AutoSynRoute \cite{AutoSynRoute} & $\checkmark$ & A & Sequence & 54.6/43.1 & 80.2/71.8 \\
       & & SCROP \cite{SCROP} & $\checkmark$ & A & Sequence & 59.0/43.7 & 78.1/65.2\\
       & & LV-Transformer \cite{LV-transformer} & $\checkmark$ & A & Sequence & NR/40.5 & NR/72.8 \\
       & & DualTF \cite{RetroEBM} & $\checkmark$ & A & Sequence & 65.7/53.6 & 84.7/74.6\\
       & & GET \cite{GET} & $\checkmark$ & A & Graph & 57.4/44.9 & 74.8/62.4\\
       & & Retroprime \cite{Retroprime} & $\checkmark$ & NA & Sequence & 64.8/51.4 & 85.0/74.0 \\
       & & GTA \cite{GTA} & $\checkmark$ & A & Graph & NR/51.1 & NR/74.8\\
       & & G2Gs \cite{Graph2Graphs} & $\times$ & A & Graph & 61.0/48.9 & 86.0/72.5\\
       & & RetroXPERT \cite{retroxpert} & $\times$ & A & Graph & 62.1/50.4 & 78.5/62.3\\
       & & Retroformer \cite{Retroformer} & $\checkmark$ & A & Sequence & 64.0/53.2 & 86.7/76.6 \\
       & & Tied-transformer \cite{tied-transformer} & $\checkmark$ & A & Sequence & NR/47.1 & NR/73.1 \\
       & & RetroLSTM \cite{Retro-LSTM} & $\checkmark$ & A & Sequence & NR/37.4 & NR/57.0\\
       & & G2GT \cite{G2GT} & $\checkmark$ & A & Graph & NR/54.1 & NR/74.5\\
       & & Stransformer \cite{stransformer} & $\checkmark$ & A & Graph & NR/43.8 & NR/NR\\
       & & GraphRetro \cite{GraphRetro} & $\times$ & A & Graph & 63.9/53.7 & 85.2/72.2\\
       & & MARS \cite{Motif-Retro} & $\checkmark$ & A & Sequence & 66.2/54.6 & 90.2/83.3\\
       \hline
    \end{tabular}
    }
    \caption{A table of recent approaches on forward and backward prediction. “End-to-end” indicates whether the model inference process is in an end-to-end manner. “Generation” indicates whether the predicted product is generated in an autoregressive or non-autoregressive manner. “Graph/Sequence” denotes the molecular representation format used in modeling. “NA” and “A” denote non-autoregressive generation and autoregressive generation respectively. We report reaction prediction results on USPTO-MIT dataset and retrosynthesis prediction results on USPTO-50K dataset. For retrosynthesis prediction model, we report both results with reaction class known and unknown. “NR” means not reported. Note that we do not put “Semi-Template-based” as an independent category.}
    \label{tab:reaction-retro}
\end{table*}

In this section, we discuss some important limitations and challenges existing in current solutions. 

\subsection{Side Products} 

\paragraph{Reaction.} Side products are missing in public USPTO benchmark datasets, which results in incomplete supervision signals. Particularly for non-autoregressive modeling, missing side products makes electron redistribution matrix  violate the conservation rule and thereby the reaction space is not fully constrained. How to complete and infer this missing information is a fundamental challenge for reaction prediction. 

\paragraph{Retrosynthesis.} Retrosynthesis directly use reaction data in USPTO and thereby all single-step analysis only contains one single outcome. Missing side products in the outcome side may not affect the general process of retrosynthesis prediction. However, side products still provide important information about leaving groups. 

\subsection{Limitations in Dataset} 

\paragraph{Reaction.} The USPTO-479K dataset has two major issues. First, reaction types are very imbalanced. Indicated by Bi \cite{NERF}, reactions with linear topology dominate reaction types while few reactions with cyclic topology exist in dataset. How to learn transferable knowledge from rare reactions is an important lesson for reaction modeling. Second, in real applications, same set of reactants can result in different products under different physical conditions, which is called multi-modality in reaction prediction. Multi-modality can provide rich information for conditional generative models to generate valid and diverse candidate products. However, most reactions in UPSTO-MIT are one-to-one mapping, which means that same set of reactants can only result in a unique major product.

\paragraph{Retrosynthesis.} The USPTO-50K dataset has two major limitations. First, the size of USPTO-50K is not large enough, which only contains 50k backward reactions. Considering many recent approaches only have slight numerical differences in top-k accuracies, the current small-scale dataset is not adequate for testing model capability. Second, the current dataset will bias edit predictions and leaving group selections. Most backward reactions only have one single edit while very few have multiple edits, which results in poor prediction accuracy in multiple-edit cases. Additionally, leaving group distribution is very imbalanced, which makes graph-based models tend to select a few frequently occurring atoms. 

%USPTO-50K is not large enough and some reaction center identification cases are hard to be differentiated due to the lack of data diversity. 

\subsection{Limitations in Evaluation} 

\paragraph{Reaction.} \cite{reactionattr} points out that the random split of USPTO-MIT is not sufficient enough for evaluations since most of the reaction types in the testing set are already covered in the training set. Therefore, harder dataset split like scaffold splits and time splits should be taken into account for future evaluations of reaction predictions. Scaffold split is testing whether the model can generalize well under out-of-distribution settings, in which the training data distribution is very different from testing data distribution. Time split is splitting reactions in the order of discovery time. This split aims at testing whether the model can truly discover new chemical reactions or not. 

\paragraph{Retrosynthesis.} Current evaluation metrics are not sufficient to evaluate retrosynthesis predictions since they require exact matching between predicted reactants and ground-truth reactants. This will lead to low diversity of generations. Although some predicted reactants are distinct from given ground-truth reactants, they are acceptable solutions for synthesis choices. Since the single-step retrosynthesis model is finally served for multi-step retrosynthetic planning purposes, low diversity for each step would make the whole planning monotonous and inflexible.  

\subsection{Challenges in Non-autoregressive Modeling} 

\paragraph{Reaction.} Non-autoregressive models \cite{NERF,reactionsink} achieve the state-of-the-art top-1 accuracy among template-free approaches, which demonstrates its high potential in reaction modeling. However, the top-k accuracies of these models are not comparable to that of autoregressive ones. These experimental results indicate that NERF has a powerful encoder-decoder architecture while its uncertainty modeling is not successful. They adopt the CVAE framework to inject uncertainty modeling. However, training CVAE will easily encounter the posterior collapse issue. In fact, the KL divergence loss term gets close to zero with few iterations during training. Also, imposing uncertainty on latent can lead to the uncontrollable generation of the following electron redistribution prediction, which will lower the validity of predicted products. To sum up, uncertainty estimation is an important challenge for non-autoregressive models. Non-autoregressive modeling with correct uncertainty estimation will revolutionize reaction prediction research. 

\paragraph{Retrosynthesis.} Non-autoregressive retrosynthesis model is very tricky since the total number of atoms in reactants must be pre-determined before attaching leaving groups. However, the distribution over the total number of reactant atoms is biased by the benchmark dataset. Therefore, to generate reactants in a non-autoregressive manner, enough blank entries should be reserved for predicted leaving groups. Unfortunately, if there are too many blank entries in the 3D adjacency matrix and atom feature matrix, the task would become a very tough imbalanced classification task since only a few blank entries will be filled up with predicted atoms and bonds. 

\subsection{Challenges in Generalization}

There are two common challenges in generalization for reaction and retrosynthesis prediction. The first challenge is out-of-distribution prediction. Out-of-distribution mainly evaluates whether the model trained on main reaction types can generalize to rare reaction types. This imbalance issue naturally exists in chemical reactions since we can expect that some reactions easily occur while some reactions rarely occur. Therefore, it is important to design stable reaction and retrosynthesis prediction models under different distribution splits. The second challenge comes from the low-level representation quality. Both problems require low-level molecular and reaction representation learning techniques. Therefore, it is significant to derive powerful molecular and reaction representations with strong generalization to related tasks. 

\section{Conclusion and Future Directions}

From the above discussions, we know that current approaches still have some major shortcomings although they have achieved promising results. Therefore, we list several future directions for further refinement of current solutions. 

\subsection{3D Molecular Information}

A chemical molecule is inherently a point cloud consisting of a set of 3D points with Cartesian coordinates. However, 3D molecular information has not been used in the literature while sequence and graph representation of molecules have been widely explored. 3D position vectors provide important complementary distance information for each pair of atoms. The relative pairwise distance can be very different in 3D Euclidean geometry compared to 2D molecular graphs. For example, atom $v_{1}$ and atom $v_{2}$ may be distant from each other in a non-euclidean molecular graph while they might be close to each other in 3D Euclidean space. This is particularly useful for reaction center ranking. Therefore, effectively incorporating 3D molecular information into modeling can facilitate the more accurate reaction and retrosynthesis predictions. 

\subsection{Diverse Benchmark Dataset and New Evaluation Metrics}

For reaction prediction, the size of the current USPTO-479K dataset is large enough. However, its modality and diversity are still not sufficient. New benchmark datasets should include more reaction types and more complex reactions. Also, different dataset splits, such as scaffold split and time split, should be included for cross-validation. For retrosynthesis prediction, the USPTO-50K dataset is at small-scale. A new benchmark dataset should be a large-scale dataset containing at least 100K samples. Furthermore, future benchmark datasets should contain more target molecules with multiple edits. In addition, a new evaluation metric for retrosynthesis is necessary and urgent. FusionRetro \cite{fusionretro} attempts to evaluate single-step retrosynthesis models in the context of multi-step planning. More diverse evaluation metrics can be designed in the future.

\subsection{Non-autoregressive Modeling} 

Just as mentioned in the previous section, the uncertainty estimation of non-autoregressive models is still inaccurate. Thus, it is crucial to explore more effective uncertainty modeling solutions for non-autoregressive models. We believe it will be a ground-breaking moment if the top-k accuracies of non-autoregressive models can be comparable to that of autoregressive ones. Non-autoregressive retrosynthesis prediction has not even been explored in current works. Due to the small size of leaving group in dataset, the significance of inference speed have been neglected in current research. We expect that the inference efficiency will be emphasized when analyzing complicated synthesis routes.

\subsection{Self-supervised Learning}

As previously discussed in limitations, reaction and retrosynthesis prediction are suffering from some issues brought by the dataset itself. Leveraging self-supervised learning (SSL) strategies on unlabeled datasets to overcome these limitations is a naturally motivated direction to explore. Although various SSL strategies have been presented for general molecular representation learning, currently seldom specific SSL strategies are designed for reaction and retrosynthesis prediction. PMSR \cite{PMSR} explores SSL strategies for retrosynthesis but it fails to achieve a strong performance. Therefore, exploring more powerful SSL strategies for both problems is a feasible direction. 

\subsection{Multi-task Learning}

Combining both tasks with other related tasks is a promising direction. DualTF \cite{RetroEBM} gives a first attempt by presenting a dual model to show that reaction and retrosynthesis prediction model can reinforce each other. FusionRetro \cite{fusionretro} and GNN-Retro \cite{GNN-Retro} leverages reaction contexts and molecular contexts respectively to improve retrosynthetic analysis. Lu and Zhang \cite{ReactionT5} show that multi-task learning can further improve reaction prediction accuracy. Intuitively, reaction classification, reaction yield prediction, and many other tasks have close connections with reaction and retrosynthesis predictions. How to leverage these tasks together is crucial for further improvements on both problems. 

\section*{Acknowledgments}
The work described here was partially supported by grants from the National Key Research and Development Program of China (No. 2018AAA0100204) and from the Research Grants Council of the Hong Kong Special Administrative Region, China (CUHK 14222922, RGC GRF, No. 2151185).

%% The file named.bst is a bibliography style file for BibTeX 0.99c
\bibliographystyle{named}
\bibliography{ijcai23}

\end{document}